\pdfoutput=1

\documentclass[11pt]{article}

\usepackage[preprint]{acl}

\usepackage{times}
\usepackage{latexsym}

\usepackage[T1]{fontenc}

\usepackage[utf8]{inputenc}

\usepackage{microtype}

\usepackage{inconsolata}

\usepackage{graphicx}

\usepackage{amsmath}
\usepackage{multirow}
\usepackage{booktabs}
\usepackage{layouts}
\usepackage{multirow}
\usepackage{threeparttable}

\usepackage[table]{xcolor}

\usepackage{tcolorbox}
\usepackage{url}
\usepackage{listings}
\usepackage{enumitem}

\usepackage[normalem]{ulem}

\definecolor{mydarkgray}{HTML}{333333}
\definecolor{mylightgray}{HTML}{F0F0F0}

\definecolor{lightgrayrule}{rgb}{0.5, 0.5, 0.5}

\definecolor{softgreen}{HTML}{66B266}
\definecolor{softred}{HTML}{E57373}

\title{Extending Audio Context for Long-Form Understanding \\in Large Audio-Language Models}

\author{
 \textbf{Yuatyong Chaichana\textsuperscript{1,2}}\thanks{Work done while interning at SCB 10X.},
 \textbf{Pittawat Taveekitworachai\textsuperscript{1}}\footnotemark[2],
 \textbf{Warit Sirichotedumrong\textsuperscript{1}}\footnotemark[2], \\
 \textbf{Potsawee Manakul\textsuperscript{1}}\footnotemark[2],
 \textbf{Kunat Pipatanakul\textsuperscript{1}}\thanks{Equal advising.}
\\
\\
 \textsuperscript{1}SCB 10X, SCBX Group,
 \textsuperscript{2}Chulalongkorn University
\\
\small{Correspondence: \texttt{6531337321@student.chula.ac.th}}
}

\begin{document}
\maketitle

\begin{abstract}

Large Audio-Language Models (LALMs) are often constrained by short audio context windows, even when their text backbones support long contexts, limiting long-form audio understanding. Prior work has introduced context-extension methods (e.g. YaRN) on unimodal LLMs, yet their application to LALMs remains unexplored. First, building on RoPE-based context extension, we introduce \textbf{Partial YaRN}, a training-free, modality-decoupled extension method that modifies only audio token positions, leaving text positions intact to preserve the base LLM’s text capabilities. Second, we propose Virtual Longform Audio Training (\textbf{VLAT}), a training strategy that extends Partial YaRN into a training-time positional augmentation. VLAT simulates diverse audio lengths during training, enabling generalization to inputs far longer than those seen in training. Our experiments on SALMONN and Qwen2-Audio confirm that Partial YaRN outperforms the original models across wide range of settings, and VLAT provides substantial performance improvement on long audio of unseen lengths.\footnote{Code and dataset are available at: \url{https://github.com/yophis/partial-yarn}.}

\end{abstract}

\begin{figure*}[t]
    \centering
    \includegraphics[width=1.0\textwidth]{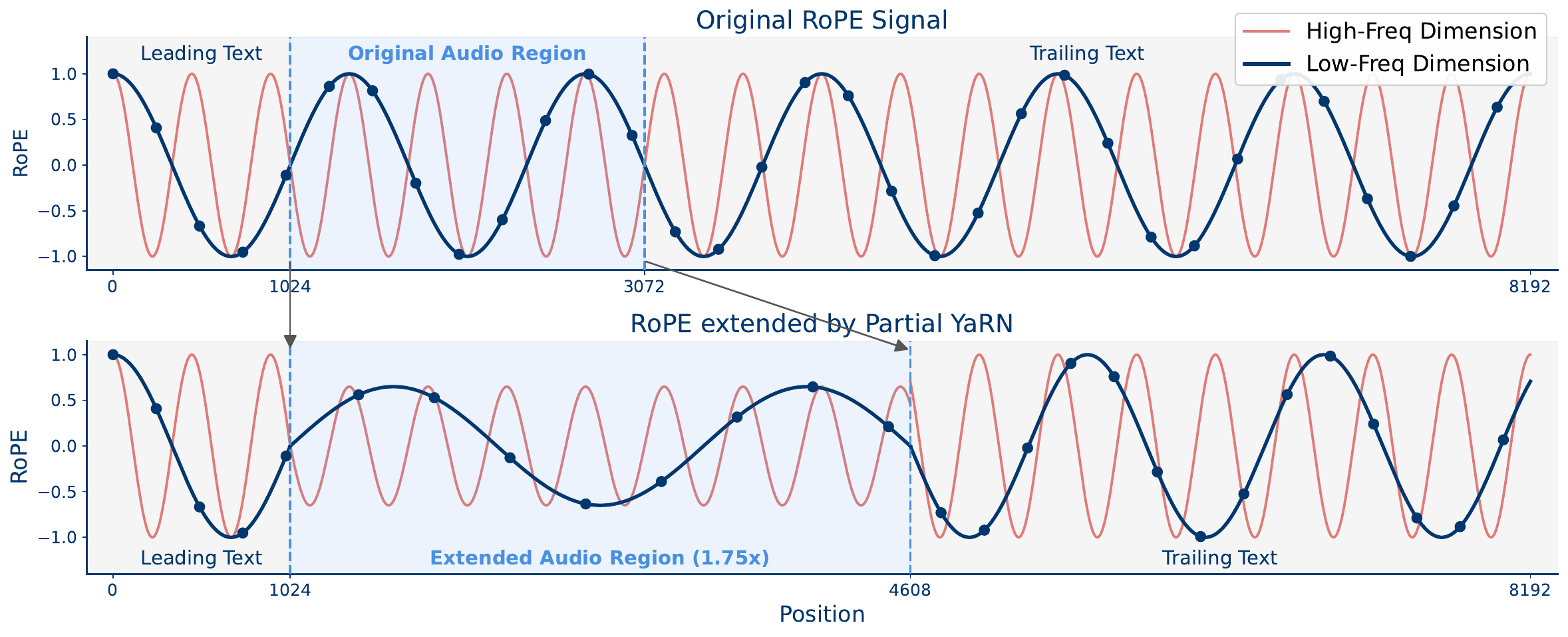}
    \caption{
        \textbf{An illustration of Partial YaRN.} The \uline{top plot} displays low (blue) and high (red) frequency dimensions of RoPE. The \uline{bottom plot} displays RoPE with Partial YaRN applied. In this example, the audio window of the low frequency dimensions are interpolated (stretched) by a factor of 1.75x to accommodate longer audio inputs, while the high frequency dimensions remain unaltered to preserve local distances and high frequency information \citep{yarn}. Within the \textit{Extended Audio Region}, the magnitudes of every dimension is scaled to serve as a proxy for attention temperature scaling. Despite that the modality-decoupled scaling can create a minor discontinuity in the signal, information from other dimensions can help the model disambiguate this.
    }

    \label{fig:rope_interpolation}
\end{figure*}

\section{Introduction}
\label{sec: Introduction}

The success of Large Language Models (LLMs) has spurred multimodal extensions, notably large audio-language models (LALMs)\footnote{Also known by other names such as Speech-aware LLMs.} that pair an audio encoder with a text backbone. By aligning audio and text into a shared representation space, LALMs can leverage the base LLM’s knowledge for complex audio understanding. However, practical use is constrained by short audio context windows: models such as SALMONN \citep{salmonn} and Qwen2-Audio \citep{qwen2audio} are typically trained on audio segments of 30s or less, and thus generalize poorly to longer inputs  \citep{fastlongspeech}. 

To address this limitation, we study the application of LLM context extension methods such as Positional Interpolation (PI) \citep{positioninterpolation} and YaRN \citep{yarn} for audio context extension in LALMs. To our knowledge, the application of these techniques for this specific, modality-bound use case has not been systematically explored. Applying such \textit{whole-context} techniques implicitly extends the audio context window as a side effect. However, as this straightforward approach alters the positional information of the entire sequence, including text tokens; it risks degrading the sophisticated language capability of the base LLM which was pretrained solely in text.

This potential drawback motivates a more targeted strategy. We introduce \textbf{Partial YaRN}, a \textit{modality-decoupled}, audio-only context extension method that modifies only the audio tokens' positional encodings. This design enables a direct comparative study against the whole-context approaches, allowing us to investigate the tradeoff between preserving position encodings of the text modality and maintaining a globally uniform positional space.

Our experiments on both training-free and fine-tuned audio context extensions reveal that extension-based methods outperform original models. However, the best choice between whole-context and modality-decoupled extensions is often model-dependent without an universally superior solution. This suggests that to build truly robust long-audio-capable models, we must address the core problem of generalization during the training process itself. Therefore, we extend Partial YaRN into a novel fine-tuning technique called \textit{Virtual Longform Audio Training} (\textbf{VLAT}). Acting as a positional augmentation technique, it simulates a wide range of audio lengths during fine-tuning, teaching the model to generalize beyond the lengths present in the training dataset. Through VLAT, we obtain substantially better generalization to longform audio of unseen lengths.

Our main contributions are as follows:
\begin{itemize}[leftmargin=*]
\setlength\itemsep{-0.3em}
    \item We propose \textit{Partial YaRN}, a training-free audio-context extension for LALMs that preserves the base language model’s text capabilities by leaving text positions unaltered.
    \item We propose \textit{Virtual Longform Audio Training (VLAT)}, a training strategy that extends Partial YaRN into a positional augmentation to simulate diverse audio lengths during training, for improved long-context generalization.
    \item We present a comparative study between existing context extension methods designed for unimodal LLMs and our proposed methods for LALMs, analyzing their performance tradeoff in both training-free and fine-tuning settings.
\end{itemize}

\section{Preliminary}
\label{sec: Preliminary}

\subsection{Rotary Positional Encoding}

As Transformer \citep{transformer} utilizes attention mechanism \citep{nmt}, which is position-invariant, explicit positional information has to be provided for the models to differentiate tokens from different positions. Our study focuses on LALMs that use Rotary Positional Encoding (RoPE) \citep{roformer}, a widely adopted positional encoding.

RoPE formulates relative positional information through the rotation of query and key vectors. Namely, given a query $\boldsymbol{q_m}$ at position $m$, and key $\boldsymbol{k_n}$ at position $n$, we have:
\begin{align*}
\boldsymbol{q}_m^\text{T} \boldsymbol{k}_n
 &= (\boldsymbol{R}_{m, \boldsymbol{\theta}} \boldsymbol{q}_{m})^\text{T} (\boldsymbol{R}_{n, \boldsymbol{\theta}} \boldsymbol{k}_n) \\
 &= \boldsymbol{q}_m^\text{T} \boldsymbol{R}_{n-m, \boldsymbol{\theta}} \boldsymbol{k}_n\,,
\end{align*}
where $\boldsymbol{\theta}$ is the set of angular frequencies (step size) of the rotation.

\subsection{Context Length Extension}
Transformers are usually unable to generalize well past the context window length they have seen during the training (e.g. 2048 for LLaMA \citep{llama}, 8192 for Mistral \citep{mistral7b}). To permit longer context length, additional fine-tuning is often required.

\paragraph{Position interpolation (PI).}
\citet{positioninterpolation} proposes that instead of extrapolating RoPE further to the region unfamiliar for the trained model, interpolation is done inside the pretrained context window. Specifically, let $L$ be the original context length of the model, and $L'$ be the target length to be extended to. Define $s = \frac{L'}{L}$ as the extension factor. PI simply downscales the base rotational frequency by the extension factor: $\frac{\theta_i}{s}$. For example, if originally each positional step is rotated by $\theta_i = 10^{\circ}$; to double the context length, the frequency (step size) can simply be halved to $5^{\circ}$. This way, the model will never have to attend to positions outside of its trained window, resulting in a more stable approach.

\paragraph{YaRN.}
\citet{yarn} identifies that interpolating every dimension equally leads to the loss of local distances and high frequency information. They propose to spread the interpolation pressure by partitioning RoPE dimensions into three frequency-based groups: (1) low frequency dimensions are interpolated, (2) high frequency dimensions are solely extrapolated without interpolation to preserve high frequency information, (3) dimensions in-between get a mix of both interpolation and extrapolation. Additionally, YaRN applies temperature scaling to the logits of attention softmax:
$$
\operatorname{AS}(\boldsymbol{Q}, \boldsymbol{K}, \boldsymbol{V}) = \operatorname{softmax}\left(\frac{\boldsymbol{Q}^T \boldsymbol{K}}{t\sqrt{d_k}}\right) \boldsymbol{V},
$$
where $t$ is the attention temperature. 

\paragraph{Other methods.}
\citet{longrope} proposes LongRoPE to account for non-uniformity across RoPE dimensions through a novel evolutionary search algorithm to determine dimension-wise interpolation factors. This method is further improved in LongRoPE2 \citep{longrope2}.

\subsection{Large Audio-language Models and Their Audio Contexts}
One broadly adopted paradigm of large audio-language models (LALMs) is the \textit{unified input space} \citep{salmonn, qwen2audio, kimi-audio, audioflamingo3} which involves transforming audio inputs into the textual input space of a base LLM. By sharing the embedding space, the strong text capability and learned knowledge of the base LLM can be leveraged while simultaneously being augmented with audio understanding ability. There also exists other LALM paradigms such as \textit{cross attention} \citep{audioflamingo, audioflamingo2} where the audio and text modalities are fused through cross-attention modules. This work focuses on models under the unified-input-space.

As with text context window in LLMs, LALMs also do not generalize well to audio with lengths much longer than the window seen during training (e.g. 7s for Pengi \citep{pengi}; 30s for Qwen2-Audio \citep{qwen2audio}).

\section{Partial YaRN}
\label{sec: Partial YaRN}

To extend existing unified input LALMs to longer audio, we hypothesize that ``\emph{LALMs already possess a sufficient general understanding of audio and text, the main bottleneck is instead their unfamiliarity to audio positions beyond the range seen during audio-text training.}''\footnote{This differs from whole-context extension in unimodal LLMs, where the total sequence length exceeds the pre-trained context window, causing an out-of-distribution problem. Here, the total sequence (audio + text) usually still fits within the original window, making the core challenge one of adapting to an unfamiliar audio length and positions, rather than extrapolating to completely unseen positions.} This hypothesis suggests a training-free solution: manipulating the backbone LLM’s positional encoding. The goal is to remap positions of a long audio input into the model’s familiar audio range.

First, we study the application of YaRN \citep{yarn} to LALMs, as an example of whole-context extension methods. Since the audio window is a part of the total context window, performing whole-context extension extends the audio window as a byproduct. However, this alters positional information for all tokens, including text, risking the degradation of the base LLM, which was pretrained solely in text. Motivated by the aforementioned hypothesis, we conceive \textit{modality-decoupled extension}, and propose \textbf{Partial YaRN}, an audio-only extension method designed to exclusively stretch the audio window. By leaving the text's positional encodings unaltered, this approach aims to extend\footnote{We use the terms \textit{extend} and \textit{stretch} interchangeably.} audio context while preserving the base model’s text capability.

\subsection{Methodology}
\label{sec: Methodology}

Inspired by PI \citep{positioninterpolation} and YaRN \citep{yarn} for whole-context extension in unimodal LLMs, we adapt the interpolation technique to LALMs in modality-decoupled manner, applying it exclusively to the audio region of the base language models.

Concretely, assuming a single audio input, let $L_\text{audio}$ be the original audio context length, $L_\text{audio}'$ be the target length, and $p$ be the position of the first audio token. We define the positional range $[p, p + L_\text{audio})$ as the \textit{original audio context window}. It can be either a predefined location for audio input, or a dynamic region enclosed by special tokens (e.g. \texttt{<speech>} and \texttt{</speech>}). We then apply our modified YaRN technique to exclusively stretch this region to the new length $L_\text{audio}'$. This creates a partially stretched positional encoding:
$$
\underbrace{[0, \; p)}_{\text{Text (unaltered)}} \oplus \; \underbrace{[p, \; p\!+\!L_\text{audio})}_{\text{\textbf{Audio} (stretched)}} \; \oplus \; \underbrace{[p\!+\!L_\text{audio}, \;  L)}_\text{Text (unaltered)}\,,
$$
where $\oplus$ denotes concatenation in the temporal axis. Note that while the final position index is denoted by $L$, the total token capacity now increases to $L + L'_{\text{audio}} - L_{\text{audio}}$ to account for the stretched audio window with higher token frequency.

Furthermore, we only partition the RoPE dimensions into two frequency-based groups (instead of three as in the original YaRN formulation): a low frequency group that undergoes pure interpolation, and a high frequency group that undergoes pure extrapolation. The rationale for this design is twofold:

\textbf{1) Consistency}: A two-group partition ensures a consistent and uniform positional encoding across the entire audio stream. YaRN's original ``in-between'' group receives a mix of interpolation and extrapolation, which would cause these RoPE frequencies to not be extended to fit the entire audio. Our approach avoids this potential distortion.

\textbf{2) Tuning Efficiency}: This simplification reduces the tuning space of the \textit{cutoff} index from two coupled variables to a single parameter.

We provide empirical validation and a more in-depth discussion of the two-group partitioning in Section~\ref{sec:Three-Group Partitioning} and Appendix~\ref{appendix: Elaboration on two-group partitioning} respectively. See Figure~\ref{fig:rope_interpolation} for an example depiction of Partial YaRN.

\subsection{Hyperparameters}
\label{sec: Hyperparameters}

Partial YaRN employs two hyperparameters:

\textbf{1) Cutoff Dimension Index}: This defines the boundary separating the RoPE dimensions into two groups. Dimensions equal or below this cutoff (low-frequency) are interpolated to stably cover longer audio context. Dimensions above it (high-frequency) are extrapolated to preserve local positional distances and high-frequency information. We default this to $0$ (interpolate every dimension).

\textbf{2) Attention Temperature}: This controls the sharpness of attention distribution within the audio context window. A higher temperature softens the distribution, preventing attention scores from collapsing to a few tokens over long sequences. We default this to $1.0$ (no temperature scaling).

Instead of directly adding a new temperature term to the attention softmax, we follow \citet{yarn} and integrate the attention temperature into magnitudes of RoPE's signals. Particularly, define $m$ and $n$ as two positions within the extended audio context window, we have the attention softmax:
\begin{align*}
\operatorname{AS}(\boldsymbol{q}_m, \boldsymbol{k}_n) = 
    \operatorname{softmax}\left(\frac{(\boldsymbol{R}_{m} \boldsymbol{q}_m)^\text{T} (\boldsymbol{R}_{n} \boldsymbol{k}_n)}{t\sqrt{d_k}}\right).
\end{align*}

Then, we invert $t$ into the two rotation matrices by scaling them with $1/\sqrt{t}$ each:
\begin{align*}
\operatorname{AS}(\boldsymbol{q}_m, \boldsymbol{k}_n) \!=\! 
    \operatorname{softmax} \! \left( \frac{\left(\frac{\boldsymbol{R}_{m}}{\sqrt{t}} \boldsymbol{q}_m \right)^\text{T} \left(\frac{\boldsymbol{R}_{n}}{\sqrt{t}} \boldsymbol{k}_n \right)}{\sqrt{d_k}} \right)\!.
\end{align*}

This reparameterization also handles attention between the unaltered text regions and the scaled audio regions. For instance, when a text query (unscaled) attends to an audio key (scaled by $1/\sqrt{t}$), the resulting logit naturally gets scaled by $1/\sqrt{t}$, creating a smoother temperature transition. Overall, we have temperature-scaled rotation matrices:
$$
\widetilde{\boldsymbol{R}}_m = \begin{cases}
\boldsymbol{R}_{\theta,m} & \text{if } m \notin [p,p+L_\text{audio}) \\
\frac{1}{\sqrt{t}} \boldsymbol{R}_{\theta/s,m} & \text{if } m \in [p,p+L_\text{audio})
\end{cases}
$$

Notice that under the default hyperparameter values, the audio-only extension reduces from YaRN to PI, we therefore specifically name Partial YaRN with default hyperparameters: \textit{Partial PI}.

\section{Experimental Setup}
\label{sec: Experimental Setting}

\subsection{Data}
Due to the limited availability of length-tailored long audio datasets, we construct custom datasets with specific audio lengths.

\paragraph{YODAS2-MCQA}
We source audio from the English subset of YODAS2 \citep{li2023yodas}. We segment audio samples into non-overlapping segments of 1, 2, 5, and 10 minutes. For each segment, we generate five multiple-choice question-answering (MCQA) pairs using Gemini 2.0 Flash \citep{gemini}, with four choices each. We design the generation prompt to ensure that each question focuses on different portions of the audio, and that they collectively cover the entire audio segment. The test set of each audio duration (1, 2, 5, and 10 minutes) has 750 QA pairs.

Additional details about dataset construction are provided in Appendix~\ref{appendix: Additional Dataset Details}.

\begin{table*}[t]
    \small
    \centering
    \begin{tabular}{llcccc}
        \toprule
        \multicolumn{2}{c}{\multirow{2}{*}{\textbf{Method}}} & \multicolumn{4}{c}{\textbf{YODAS2-MCQA}} \\
        \cmidrule(lr){3-6}
        \multicolumn{2}{c}{} & 1 min & 2 mins & 5 mins & 10 mins \\
        \midrule

        \rowcolor[gray]{0.95}
        \multicolumn{2}{l}{\textbf{GPT 4o}} & 89.10 & 92.91 & 90.40 & 83.65 \\
        \rowcolor[gray]{0.95}
        \multicolumn{2}{l}{\textbf{Gemini 2.0 Flash}} & 94.70 & 95.86 & 92.80 & 89.87 \\
        \midrule

        \rowcolor[gray]{0.95}
        \multicolumn{2}{l}{\textbf{SALMONN}} & \multicolumn{4}{l}{} \\
        & Vanilla & 49.01 & 46.13 & 32.93 & 23.47 \\
        \multicolumn{6}{l}{\textit{Stretching from 30s (original context)}} \\
        & Whole PI & 39.87 \textcolor{softred}{($-9.14$)} & 5.73 \textcolor{softred}{($-40.40$)} & 0.13 \textcolor{softred}{($-32.80$)} & 23.20 \textcolor{softred}{($-0.27$)} \\
        & Partial PI & 50.07 \textcolor{softgreen}{($+1.06$)} & 45.60 \textcolor{softred}{($-0.53$)} & 12.40 \textcolor{softred}{($-20.53$)} & 1.20 \textcolor{softred}{($-22.27$)} \\
        & Whole YaRN & \underline{54.70} \textcolor{softgreen}{($+5.69$)} & \textbf{59.87} \textcolor{softgreen}{($+13.74$)} & \textbf{47.20} \textcolor{softgreen}{($+14.27$)} & 30.13 \textcolor{softgreen}{($+6.66$)} \\
        & Partial YaRN & \textbf{57.35} \textcolor{softgreen}{($+8.34$)} & \underline{59.60} \textcolor{softgreen}{($+13.47$)} & 38.53 \textcolor{softgreen}{($+5.60$)} & 32.93 \textcolor{softgreen}{($+9.46$)} \\
        \multicolumn{6}{l}{\textit{Stretching from 2 mins (observed context)}} \\
        & Whole PI & \textit{\textcolor{gray}{49.01}} & \textit{\textcolor{gray}{46.13}} & 13.07 \textcolor{softred}{($-19.86$)} & 23.60 \textcolor{softgreen}{($+0.13$)} \\
        & Partial PI & \textit{\textcolor{gray}{49.01}} & \textit{\textcolor{gray}{46.13}} & 28.27 \textcolor{softred}{($-4.66$)} & 4.13 \textcolor{softred}{($-19.34$)} \\
        & Whole YaRN & \textit{\textcolor{gray}{49.01}} & \textit{\textcolor{gray}{46.13}} & \underline{45.87} \textcolor{softgreen}{($+12.94$)} & \textbf{38.27} \textcolor{softgreen}{($+14.80$)} \\
        & Partial YaRN & \textit{\textcolor{gray}{49.01}} & \textit{\textcolor{gray}{46.13}} & 41.47 \textcolor{softgreen}{($+8.54$)} & \underline{33.45} \textcolor{softgreen}{($+9.98$)} \\
        \midrule

        \rowcolor[gray]{0.95}
        \multicolumn{2}{l}{\textbf{Qwen2-Audio 7B Instruct}} & \multicolumn{4}{l}{} \\
        & Vanilla & 72.71 & \textbf{75.87} & 55.33 & 22.00 \\
        \multicolumn{6}{l}{\textit{Stretching from 30s (original context)}} \\
        & Whole PI & 64.50 \textcolor{softred}{($-8.21$)} & 7.73 \textcolor{softred}{($-68.14$)} & 5.87 \textcolor{softred}{($-49.46$)} & 8.67 \textcolor{softred}{($-13.33$)} \\
        & Partial PI & 72.45 \textcolor{softred}{($-0.26$)} & 67.73 \textcolor{softred}{($-8.14$)} & 40.13 \textcolor{softred}{($-15.20$)} & 18.40 \textcolor{softred}{($-3.60$)} \\
        & Whole YaRN & \textbf{73.11} \textcolor{softgreen}{($+0.40$)} & \underline{74.27} \textcolor{softred}{($-1.60$)} & 50.93 \textcolor{softred}{($-4.40$)} & 31.07 \textcolor{softgreen}{($+9.07$)} \\
        & Partial YaRN & \underline{72.72} \textcolor{softgreen}{($+0.01$)} & 73.60 \textcolor{softred}{($-2.27$)} & 48.53 \textcolor{softred}{($-6.80$)} & 28.53 \textcolor{softgreen}{($+6.53$)} \\
        \multicolumn{6}{l}{\textit{Stretching from 2 mins (observed context)}} \\
        & Whole PI & \textit{\textcolor{gray}{72.71}} & \textit{\textcolor{gray}{75.87}} & 46.93 \textcolor{softred}{($-8.40$)} & 21.20 \textcolor{softred}{($-0.80$)} \\
        & Partial PI & \textit{\textcolor{gray}{72.71}} & \textit{\textcolor{gray}{75.87}} & 59.60 \textcolor{softgreen}{($+4.27$)} & 45.73 \textcolor{softgreen}{($+23.73$)} \\
        & Whole YaRN & \textit{\textcolor{gray}{72.71}} & \textit{\textcolor{gray}{75.87}} & \underline{60.27} \textcolor{softgreen}{($+4.94$)} & \underline{47.60} \textcolor{softgreen}{($+25.60$)} \\
        & Partial YaRN & \textit{\textcolor{gray}{72.71}} & \textit{\textcolor{gray}{75.87}} & \textbf{60.40} \textcolor{softgreen}{($+5.07$)} & \textbf{48.00} \textcolor{softgreen}{($+26.00$)} \\
        \bottomrule

    \end{tabular}
    \caption{\textbf{Training-free audio context extension performance (Accuracy).} We compare extension methods under two strategies: stretching from the original 30s context and the empirically observed 2mins context. We see that stretching from the observed 2mins context proves significantly more effective. Comparing between the two top approaches: Whole and Partial YaRN, we find their performance to be closed, suggesting that preserving the positional information of text tokens might not be of utmost importance, at least in this MCQA setting of ours.}
    \label{tab:performance_zeroshot}
\end{table*}

\subsection{Models}
\label{sec: Model and Audio Processing}

We select two widely-used open-weights LALMs: SALMONN \citep{salmonn} and Qwen2-Audio \citep{qwen2audio}. Both models utilize Whisper-encoder \citep{whisper} as audio encoder, which is limited to processing audio of up to 30s. We handle longer audio inputs by segmenting them into non-overlapping 30s chunks, where each chunk is encoded independently. 

The number of audio representations differs considerably between these models. For each 30s chunk on 50Hz Whisper-encoder, SALMONN generates a sequence of 88 audio tokens (with Q-Former), whereas Qwen2-Audio produces a sequence of 750 tokens (with a 2x downsampling).

\subsection{Baseline Methods}
We evaluate Partial PI (default hyperparameters) and Partial YaRN (tuned hyperparameters) against the following methods:

\textbf{1) Vanilla}: Unmodified base models. Long audio inputs are passed directly without any manipulation to their RoPE.

\textbf{2) Whole Position Interpolation (PI)} \citep{positioninterpolation}: Every RoPE dimension is uniformly interpolated across the whole context window. This serves as the primitive whole-context baseline.

\textbf{3) Whole YaRN} \citep{yarn}: Frequency-based RoPE interpolation with attention scaling, applied to the whole context window. This serves as the primary whole-context baseline.

These three baselines require no tuning as Vanilla and Whole PI don't utilize any hyperparameter, and YaRN has predefined cutoffs and a closed-form formula for temperature.\footnote{Unlike Whole YaRN for unimodal LLMs, where information per token is relatively consistent, compression of audio tokens varied significantly across different LALMs as mentioned. This likely influences the optimal hyperparameter choice considerably, as discussed in Appendix~\ref{Appendix:Hyperparameter Sensitivity of Partial YaRN}.}

\paragraph{Reference models.}
We also report performance of two proprietary models capable of long audio context: GPT-4o \citep{gpt4o} and Gemini 2.0 Flash \citep{gemini}, in order to provide a broader perspective of LALMs and to validate the quality of our dataset.\footnote{The model versions are \texttt{gpt-4o-mini-2024-07-18} and \texttt{gemini-2.0-flash-001} respectively.}

\begin{table*}[t]
    \setlength\tabcolsep{4pt}
    \small
    \centering

    \begin{tabular}{lcccccc}
        \toprule
        & \multicolumn{3}{c}{\textbf{Vanilla Inference}} & \multicolumn{3}{c}{\textbf{Partial PI Inference}} \\
        \cmidrule(lr){2-4} \cmidrule(lr){5-7}
        \textbf{Training Method} & 2 mins & 5 mins & 10 mins & 2 mins & 5 mins & 10 mins \\
        \midrule
        Vanilla Fine-tuning & 97.60 & 89.07 & 32.76 & 96.40 & 89.78 & 75.56 \\
        Virtual Longform & 97.60 \textcolor{softgreen}{($+0.00$)} & \textbf{91.42} \textcolor{softgreen}{($+2.35$)} & \textbf{75.11} \textcolor{softgreen}{($+42.35$)} & \textbf{96.87} \textcolor{softgreen}{($+0.47$)} & \textbf{91.91} \textcolor{softgreen}{($+2.13$)} & \textbf{81.73} \textcolor{softgreen}{($+6.17$)} \\
        \bottomrule
    \end{tabular}

    \caption{\textbf{Virtual Longform Audio Training (VLAT) enables strong generalization to unseen audio lengths (Accuracy).} We compare standard Vanilla fine-tuning and our proposed VLAT on Qwen2-Audio, evaluated with (\textit{Partial PI Inference} column group) and without (\textit{Vanilla Inference} column group) Partial PI during inference.}
    \label{tab:virtual longform}
\end{table*}

\section{Main Results}
\label{sec: Main Results}

In this section, we comparatively study the effect of whole-context and modality-decoupled context extensions on both the training-free and fine-tuning settings for audio segments up to 10mins.

\subsection{Training-free Audio Context Extension}
\label{sec: Training-free audio context extension}

First, we extend the audio context length without training. We report the results in Table~\ref{tab:performance_zeroshot}, showing that: no single method---whole-context or modality-decoupled---is universally superior. Instead, the performance varies between the different models and the degree of extension. More importantly, we observe that \textit{both models generally retain consistent performance up to 2mins}, suggesting that their innate audio context lengths are longer than just 30s, possibly a result from multi-audio training.

\paragraph{Extension of SALMONN.}
When stretching from the 30s window, extension methods provide substantial performance gains compared to the Vanilla baseline. We observe that Partial YaRN shows the strongest performance on the 1min and 10mins settings, achieving 57.35\% and 32.93\% accuracies. On 2mins, both Whole and Partial YaRN perform similarly. However, Whole YaRN outperforms Partial YaRN at 5mins by a large margin. Possible reasons for this performance flip are (1) higher expressivity and less compression pressure from Whole YaRN's additional ``in-between'' dimension group, outweighs the preservation of the base language capability in this 5mins setting, or (2) noise in the hyperparameter tuning process of Partial YaRN, which is an inherent drawback.

\paragraph{Extension of Qwen2-Audio.}
In contrast to SALMONN, Qwen2-Audio demonstrates different characteristics. The Vanilla baseline is more robust, outperforming all stretch-from-30s methods at 2mins. Both Whole and Partial YaRN struggle to provide significant gains and even underperform the baseline at moderate lengths (2 and 5mins) when stretching from the original 30s context; however, they manage to attain considerable gains on the 10mins setting. This indicates that Qwen2-Audio likely possesses a strong audio-length generalization capability or longer innate audio context window. For this model, simply applying an extension method from the base context does not guarantee an improvement. This suggests that the model's high intrinsic audio-length generalizability is superior to the extension methods that inevitably cram the positional information as a side effect.

\paragraph{Extending from observed audio context.}
As mentioned in the beginning of this section that we can observe both model's innate audio contexts to be around 2mins instead of just 30s, here we experiment with extending from this \textit{observed audio context of 2mins}, to the target lengths. For both SALMONN and Qwen2-Audio, performance on the long audio improves substantially when the interpolation is anchored from the 2mins context instead of the original 30s. Notably on Qwen2-Audio, this strategy boosts Partial YaRN's performance at 10mins from 28.53\% to a 48.00\%, an absolute improvement of almost 20\%, yielding a final performance 26\% higher than the Vanilla baseline. This finding suggests that determining the extension by observing the vanilla performance may be a more critical factor than the choice between whole-context or modality-decoupled methods, especially on large extension ratios.

\subsection{Fine-tuning for Audio Context Extension}
\label{sec: fine-tuning for audio context extension}
Next, we study training-based extension methods. Keeping other components frozen, we fine-tune the base language model of Qwen2-Audio using a \textit{single epoch} LoRA \citep{lora} of rank 8 on the YODAS2-MCQA training set of the corresponding length. Two LoRA settings are explored: adapting only the query projection (q), and adapting query, key, value, and output projections (qkvo). Due to the high cost of tuning Partial YaRN's hyperparameters during training, we only use the default configuration (Partial PI) for the audio-only method. Whole PI is omitted due to its poor performance as previously observed. 

The results are presented in Table~\ref{tab:fine-tuning_results}, showing that fine-tuning with a context extension method dramatically outperforms the vanilla fine-tuning baseline, especially at longer audio lengths. In the qkvo setting, both Whole YaRN (83.47\%) and Partial PI (83.07\%) achieve an absolute performance gain of around 19\% over the Vanilla baseline (64.93\%) at 10mins. This highlights the benefit of integrating a context extension method into the fine-tuning process for better context extension.

Comparing the whole-context and modality-decoupled approaches, we find that their performance is generally competitive. While Whole YaRN triumps when only adapting the query projection weights, suggesting its ease of adaptation. Their performance converges under standard LoRA practice which adapts qkvo. This result confirms that both whole-context and audio-only methods are highly effective strategies for fine-tuning.

\begin{table}[h]
    \small
    \centering

    \begin{tabular}{llccc}
        \toprule
        \textbf{Adapt} & \textbf{Method} & \textbf{2 mins} & \textbf{5 mins} & \textbf{10 mins} \\
        \midrule
        \multirow{3}{*}{\textbf{q}} & Vanilla & 95.07 & 78.93 & 21.60 \\
                                     & Whole YaRN & \textbf{96.53} & \textbf{88.53} & \textbf{73.73} \\
                                     & Partial PI & 96.13 & 83.73 & 69.20 \\
        \cmidrule(lr){1-5}
        \multirow{3}{*}{\textbf{qkvo}} & Vanilla & 96.13 & 85.20 & 64.93 \\
                                      & Whole YaRN & 97.47 & \textbf{93.60} & \textbf{83.47} \\
                                      & Partial PI & \textbf{98.00} & 93.33 & 83.07 \\
        \bottomrule
    \end{tabular}

    \caption{\textbf{Fine-tuned audio context extension performance (Accuracy).} We fine-tune Qwen2-Audio on YODAS2-MCQA, under two LoRA settings: adapting only \textit{q}, and adapting \textit{qkvo}. Each model is fine-tuned and evaluated using the same extension method.}
    \vspace{-1em}
    \label{tab:fine-tuning_results}
\end{table}

\section{Virtual Longform Audio Training: Generalizing Beyond the Training Data}
\label{sec: Virtual Long Range}

Our previous results show that while the training-free methods are beneficial, their effectiveness is often model and length dependent. Based on this, we turn to address the core generalization problem during the training process, and propose to fine-tune LALMs with Partial YaRN repurposed as a positional augmentation technique we call ``\textit{Virtual Longform Audio Training}'' (VLAT).

\subsection{Methodology}
The goal of VLAT is to simulate and expose models to audio context windows of diverse lengths during training. Let $L_\text{data}$ be the actual length of a training audio sample. For each sample, we obtain a ``virtual'' source length, $L_\text{virt}$, as the model's default audio context length \textit{times} a factor randomly sampled from the range [1, 5, 10, 15, 20, 25]x. We then apply Partial YaRN\footnote{Whole YaRN can't be used here, as we observe that it diverges quickly when training under the VLAT framework.} to stretch or compress the $L_\text{virt}$-long positional window to $L_\text{data}$.

For instance, consider an audio sample with $L_\text{data}=2$mins, and we draw a $L_\text{virt}$ of 10mins, Partial YaRN will compress a 10mins context window to 2mins for this audio. This process effectively simulates the 2mins audio to be 10mins long virtually, thereby familiarizing the model to longer audio context and improving its ability to generalize to genuinely long audio at inference time.

Randomizing positional encoding has previously been explored in \citet{randomizedpositionalencoding}, where positional indices are randomly downsampled to let the model see larger positional values. Rather than using $\{1,2,\dots,N\}$ for an length-$N$ input, they use a randomly subsampled values such as $\{1,4,\dots,L\}$ with $L > N$ instead. In contrast, VLAT differs in employing interpolation inside the context window to create a dense, \textit{continuous} space of positions, unlike sparse integer subsampling. It is also bidirectional, teaching the model through both compressed and stretched contexts. Furthermore, VLAT is a targeted, modality-decoupled method that modifies only the audio tokens, preserving the base LLM's original textual space.

\subsection{Evaluation}
We fine-tune Qwen2-Audio on the 2mins YODAS2-MCQA training set\footnote{To prevent data leakage, the train/test splits were performed at the video level, ensuring no source material overlaps between the training set and any of the test sets.} using two methods: a standard Vanilla fine-tuning, and VLAT. We employ 10-epoch qkvo LoRA. We then evaluate the models from both training methods under two different extensions at inference: Vanilla and Partial PI. Results are averaged from 3 random seeds. See Appendix~\ref{Appendix: Different VLAT's Virtual Length Sampling Strategies} for detail on VLAT configuration.

\subsection{Results}

As shown in Table~\ref{tab:virtual longform}, VLAT greatly improves generalization to unseen audio lengths. When evaluated with vanilla inference, the model trained with Virtual Longform shows a dramatic improvement on 10mins audio, increasing accuracy from 32.76\% to 75.11\%, closing in on the previous 1-epoch direct-fine-tuning result. This highlights the effectiveness of the training technique for training LALMs to generalize far beyond the audio lengths present in their training data, mitigating a key bottleneck in the development of robust long-audio models.

Furthermore, as can be seen in "Partial PI Inference" columns, VLAT's benefit is complementary to inference-time extension. When Partial PI is applied during evaluation, the VLAT-trained model again outperforms its vanilla-trained counterpart, achieving the highest overall 10mins performance of 81.73\%. This indicates that VLAT and inference-time extension are compatible strategies, and their combination yields the most robust long-audio-context performance.

We show in Appendix~\ref{Appendix: Different VLAT's Virtual Length Sampling Strategies} that VLAT's performance can be even more enhanced by using a narrower range for virtual length $L_\text{virt}$.

\section{Analysis}
\label{sec: Analysis}

\subsection{Ablation Study}
\label{sec: Ablation study}
Here, we conduct an ablation study to isolate and analyze the individual contributions of the two key components in Whole YaRN \citep{yarn} and Partial YaRN: (1) frequency grouping and (2) attention temperature scaling. We note that by removing both of these components, the methods converge to Whole PI \citep{positioninterpolation} and Partial PI. The results are presented in Table~\ref{tab:ablation}.

\begin{table}[h]
    \small
    \setlength\tabcolsep{3.5pt}
    \centering

    \begin{tabular}{lcccc}
        \toprule
        & \multicolumn{2}{c}{\textbf{Partial}} & \multicolumn{2}{c}{\textbf{Whole}} \\
        \cmidrule(lr){2-3} \cmidrule(l){4-5}
        \textbf{Method} & 2 mins & 10 mins & 2 mins & 10 mins \\
        \midrule
        \rowcolor[gray]{0.95}
        \multicolumn{5}{l}{\textbf{SALMONN}} \\
        \quad YaRN & \textbf{59.60} & \textbf{32.93} & \textbf{59.87} & \textbf{30.13} \\
        \quad - Freq grouping & 56.93 & 4.53 & 23.33 & 29.20 \\
        \quad - Attn temp & 48.13 & 23.87 & 41.07 & 23.20 \\
        \quad PI & 45.60 & 1.20 & 5.37 & 23.20 \\
        \arrayrulecolor{black!70}
        \cmidrule(l){1-5}
        \arrayrulecolor{black}
        \rowcolor[gray]{0.95}
        \multicolumn{5}{l}{\textbf{Qwen2-Audio}} \\
        \quad YaRN & \textbf{73.60} & \textbf{28.53} & 73.47 & \textbf{25.33} \\
        \quad - Freq grouping & 66.00 & 28.13 & 17.07 & 9.87  \\
        \quad - Attn temp & 73.47 & 18.80 & \textbf{75.07} & 22.80 \\
        \quad PI & 67.73 & 18.40 & 7.73 & 8.67 \\
        \bottomrule
    \end{tabular}

    \caption{\textbf{Ablation of frequency grouping and attention temperature components (Accuracy).} Removing either the frequency grouping or the attention temperature leads to a large drop in performance.}
    \label{tab:ablation}
\end{table}

We can see that removing either the frequency grouping or the attention temperature generally leads to a large drop in performance, especially on higher extension ratios. Both the frequency partitioning and attention temperature scaling are crucial for robustly extending the audio context length of LALMs. Their combined application within the Whole and Partial YaRN frameworks generally yield the most effective and stable performance.

\subsection{Partial YaRN with Three-group Frequency Partitioning}
\label{sec:Three-Group Partitioning}

For the Partial YaRN method, we modify the original 3-group frequency partition from original YaRN to a 2-group method as discussed in Section~\ref{sec: Methodology} and Appendix~\ref{appendix: Elaboration on two-group partitioning}. Here we conduct a direct empirical comparison between the 2-group and 3-group approaches.

For the 3-group baseline, we implement Partial YaRN using YaRN's original partitioning scheme and its recommended hyperparameters. We compare this against our proposed 2-group Partial YaRN using the hyperparameters from our main experiments (Section~\ref{sec: Training-free audio context extension}). We present the results in Table~\ref{tab:2v3_group}.

\begin{table}[h]
    \setlength\tabcolsep{3.5pt}
    \small
    \centering
    \begin{tabular}{lcccc}
        \toprule

        \multicolumn{2}{c}{} & \multicolumn{3}{c}{\textbf{YODAS2-MCQA}} \\
        \cmidrule(lr){3-5}
    
        \textbf{Model} & \textbf{Freq Grouping} & {2 mins} & {5 mins} & {10 mins} \\
        \midrule
        \multirow{2}{*}{SALMONN} & 2-Group& \textbf{59.60} & 38.53 & \textbf{32.93} \\
        & 3-Group & 56.00 & \textbf{41.87} & 25.46 \\
        \midrule
        \multirow{2}{*}{Qwen2-Audio} & {2-Group} & \textbf{73.60} & \textbf{48.53} & \textbf{28.53} \\
        & 3-Group & 72.00 & 20.00 & 16.53 \\
        \bottomrule
    \end{tabular}
    
    \caption{\textbf{Performance Comparison of Two-Group vs. Three-Group Partitioning (Accuracy).} We compare our Partial YaRN with 2-group frequency partitioning used throughout the work, against a 3-group version.}
    \label{tab:2v3_group}
\end{table}

We see that our 2-group partitioning broadly outperform 3-group in our modality-specific scenario. This is most evident with the Qwen2-Audio model, where the 3-group method suffers a large performance drop on 5mins and 10mins audio, with an accuracy gap of 28.53\% and 12.00\% respectively. Overall, this result validates the effectiveness and reliability of our proposed 2-group scheme.

\section{Related Work}
\label{sec: Related Works}

This section discusses additional related works that aren't addressed in Section~\ref{sec: Preliminary}.

\paragraph{Positional encodings.}
Absolute positional encoding was originally proposed in \citet{transformer}. It utilizes sinusoidal waves to uniquely represent absolute position of each token. Later, \citep{bert, albert, electra} use trainable vectors for representing absolute positions, allowing the models to learn the positional encoding for themselves. However, it cannot extrapolate to sequences longer than the maximum length it was trained on. \citet{relativepositionalencoding} encodes the relative distances between the tokens instead of absolute positions. \citet{transformerxl} introduces additional trainable position-agnostic components to distinguish between content-based and location-based attention behaviors. \citet{t5} groups the relative distances into buckets, and associate each bucket with a learnable scalar for adding to attention logits. TUPE \citep{tupe} uses separate linear projections for the token and positional encodings. ALiBi \citep{alibi} simply adds linear biases with a learnable slope into the attention scores, imposing a strong locality preference. Several later works extend RoPE for multimodal modeling, such as Multimodal RoPE \cite{qwen2.5vl}, VideoRoPE \citep{videorope}, and VRoPE \citep{vrope}.

\paragraph{Large video-language models.}
Video input is similar to the audio in that they're both naturally continuous and can be of arbitrary lengths. Many Large Video-Language Models (LVLMs) \citep{video-llama, video-gpt, videollava, chatunivi, timechat}, adopt architectural paradigm similar to the unified input space LALMs we studied, where video frames are embedded and projected into a base LLM's input space. However, the context length problem is even more exacerbated in the video domain due to the high dimensionality and large number of tokens needed to represent even short videos. To improve efficiency, the LVLM community has largely focused on input compression techniques like query aggregation \citep{video-llama, videochat, moviechat}, frame pooling \citep{valley, video-gpt}, and feature merging \citep{longvlm}. These strategies primarily aim to reduce the number of tokens before they enter the base LLM. Our work offers a complementary perspective: instead of altering the input, we manipulate the positional space within the model to accommodate a longer sequence, similar to \citet{zhang2025long} in LVLMs. The principles of Partial YaRN and VLAT are likely applicable to the video domain as well, hopefully providing a distinct path to achieve long-video-context understanding and generative control in synergy with existing methods.

\paragraph{Long-audio LALMs.}
Concurrent works have proposed LALMs that have inherently long audio-context. Audio Flamingo 3 \citep{audioflamingo3} is capable of processing audio up to 10mins due long-audio training. Voxtral \citep{voxtral} can handle up to 40mins due to the long 32k context of its LLM backbone. FastLongSpeech \citep{fastlongspeech} proposes Iterative Fusion which incrementally merges the audio representation based on their information density. This compressed representation facilitates efficient processing of long audio. Qwen3-Omni \citep{qwen3omni} is a multimodal model that can accept text, video, and audio as input. It employs a custom Audio Transformer which is trained in an encoder-decoder fashion similar to the Whisper model. Qwen3-Omni is also capable of processing audio of up to 40mins.

While these models achieve impressive long-audio performance, they do so by designing new architectures, curating large volumes of long-audio data, and undertaking extensive training. Our work investigates an orthogonal yet complementary direction of efficiently extending the context of existing LALMs. Rather than requiring costly data acquisition and retraining, we provide a lightweight, architecture-agnostic "drop-in" enhancement (Partial YaRN) and an efficient fine-tuning strategy (VLAT), offering a practical and accessible approaches for improving the long-audio capabilities of a wide range of existing models.

\section{Conclusion and Future Work}
This work addressed the challenge of extending audio context windows of LALMs by studying the application of LLM's whole-context extension methods, and proposing a training-free audio-only extension Partial YaRN and the VLAT training strategy. We demonstrated the effectiveness of extension-based methods over non-extension baselines across various long-audio settings. Leveraging our finding that SALMONN and Qwen2-Audio retain consistent performance up to 2mins audio window, we achieved an even stronger performance by extending from the 2mins window instead of 30s. Later, we showed that fine-tuning through VLAT allows the models to generalize to audio contexts far exceeding the lengths of the training data, offering a robust and data-efficient pathway to develop LALMs with better longform understanding.

Future work could apply Partial YaRN and VLAT to vision models. Beyond content understanding, these approaches could also be explored for controllable generation in multimodal models, such as multi-scale image synthesis, or even more finegrained decoupling for non-uniform speed control throughout audio or video generation.

\section*{Limitations}

\paragraph{Evaluation scope.}
Our experiments are conducted exclusively on the multiple-choice question answering tasks. We made this choice to ensure a precise and unambiguous measurement of the model performance under different audio lengths. This mitigates the confounding variables inherent in evaluating open-ended generation, such as judging the semantic equivalence or stylistic alignment. However, a key tradeoff of this focused evaluation is its limited ability to assess the nuanced generative and language modeling capabilities of the base LLM. Consequently, while our results validate the model's ability to retrieve information from audio, they do not fully test the importance of text preservation of the audio-only methods like Partial YaRN. Furthermore, our study does not assess the performance across other long-audio task formats found in benchmarks such as \citet{fastlongspeech}, \citet{blab}, or \citet{audioflamingo3}.

\paragraph{Reliance on synthetic data.}
Our study utilizes YODAS2-MCQA, a custom dataset synthesized from the YODAS2 corpus using Gemini 2.0 Flash. While we carefully steered the synthesis process and validated a subset of the data, it's still possible for the synthesized samples to carry noise, repetition, and hallucinations that may affect the fidelity of our performance estimates compared to real-world scenarios.

\paragraph{Hyperparameter tuning burden.}
Unlike the original YaRN which has generally good predefined cutoff dimensions and a closed-form formula for determining the attention temperature, our proposed Partial YaRN requires tuning of two hyperparameters, thereby introducing additional computational burden.

\section*{Acknowledgments}
YC would like to thank Ekapol Chuangsuwanich for insightful discussion on the earlier version of this work. He also thanks Thanapat Trachu for valuable guidance on the illustration of Partial YaRN, and Konpat Preechakul for illuminating discussion on extensions of this work to the vision domain. We gratefully acknowledge the compute resources provided by Hatari NEXT, on which the experiments for this research were run.


\nocite{audiogpt, speechgpt, audiopalm, speechllama, moshi, pengi, ltu, ltu-as, qwenaudio, gama, typhoon-audio, step-audio, baichuan-audio} 

\bibliography{custom}
\clearpage

\appendix

\section{Elaboration on Two-group Partitioning of Partial YaRN}
\label{appendix: Elaboration on two-group partitioning}

The decision to simplify the original three-group frequency partition of YaRN \citep{yarn} into a two-group system for Partial YaRN is a deliberate design choice aimed at ensuring positional consistency across the entire extended audio stream. The original YaRN formulation partitions RoPE dimensions into: 1) a low-frequency group that is purely interpolated, 2) a high-frequency group that is purely extrapolated, and 3) an "in-between" group that receives a mix of both, with a ramped interpolation factor.

While effective for whole-context extension, this "in-between" group becomes problematic in our partial, audio-only extension scenario. Consider a 2mins audio input, which requires a 4x extension of the model's 30s original audio context.
\begin{itemize}
\item The high-frequency dimensions, when extrapolated (a 1x factor), correctly preserve local relative distances, which is desirable.
\item The low-frequency dimensions, responsible for quasi-absolute positioning, are fully interpolated by a 4x factor. This is crucial as it maps the entire 2mins audio into the model's familiar positional range (original audio context).
\item The "in-between" dimensions, however, would receive interpolation factors between the 1x and 4x (e.g. 2x). This means that for these specific dimensions, the model's original context window is stretched to cover only the first minute of the 2mins audio. The second minute would lie in an extrapolated, out-of-distribution region.
\end{itemize}

This dimensional inconsistency risks creating a biased or distorted representation of the latter half of the audio. The model would receive conflicting signals about whether a token is "inside" or "outside" its familiar context.

By simplifying to a two-group partition, we circumvent this issue entirely. Every dimension is forced into a binary choice: either it is fully interpolated to cover the entire audio duration (low-frequency), or it is fully extrapolated to preserve local structure (high-frequency). This ensures that the model maintains a consistent positional understanding across all dimensions for the entirety of the audio input, avoiding the representational bias that partial coverage would introduce.

See Section~\ref{sec:Three-Group Partitioning} for direct performance comparison between 2 and 3-group frequency partitioning versions of Partial YaRN.

\section{Implementation of Partial YaRN via \texttt{position\_ids} Manipulation}
\label{appendix:implementation_details}

A key implementation detail of our work is how we apply the positional stretching to only a subsection of the input sequence. While the original YaRN implementation modifies RoPE's frequencies ($\boldsymbol{\theta}$), this approach isn't straightforward for a \textit{partial} modification. Instead, our implementation of Partial YaRN directly manipulates the \texttt{position\_ids} and multiply it with the unmodified frequencies. This section details the implementation our method.

\paragraph{Modifying positional indices.}

In most Transformer architectures, including the models used in this work, the RoPE mechanism computes embeddings based on an input tensor called \texttt{position\_ids}. This tensor is simply a sequence of integers representing the absolute position of each token (i.e. \texttt{[0, 1, 2, ..., N-1]}). Our method leverages this by modifying a section this tensor. The process is as follows:

\begin{enumerate}
    \item \textbf{Identify the Audio Region:} First, we identify the start and end indices of the audio tokens within the \texttt{position\_ids} tensor. This creates three distinct segments: the leading text positions (unaltered), the audio positions (to be stretched), and the trailing text positions (unaltered relative to each other).

    \item \textbf{Stretch the Audio Positions:} We replace the original sequence of integer positions for the audio segment with a new sequence generated via \texttt{torch.linspace}. For example, if the original audio positions were \texttt{[p, p+1, ..., p+base\_L\_audio-1]} and the new audio input has \texttt{L\_audio} tokens, we generate \texttt{L\_audio} new positions by interpolating within the original range: \texttt{torch.linspace(start=p, end=p+L\_audio-1, steps=L\_audio)}. This effectively ``stretches'' the familiar positional space to accommodate the longer audio input.

    \item \textbf{Concatenate:} The final \texttt{position\_ids} tensor is constructed by concatenating the leading text positions, the new stretched audio positions, and the trailing text positions. This new \texttt{position\_ids} is now ready to be use for subsequent construction of partially stretch RoPE signal for the relevant dimensions.
\end{enumerate}

\paragraph{Computational overhead.}
This implementation introduces negligible computational overhead in practice. The only additional computations are incurred during the construction of the \texttt{position\_ids} tensor, which involves a few tensor slicing operations, a single \texttt{torch.linspace} call, and a final concatenation. These are highly optimized, vectorized operations that constitute a minuscule fraction of the total computation in a full forward pass. Crucially, this precomputation is a \uline{one-time cost} per input sample, applied only during the initial processing of the prompt that contains the long audio. During the subsequent autoregressive generation steps, where tokens are generated one at a time, our method adds no overhead whatsoever. The model reverts to the standard process of incremental positional stepping, just as it would without any context extension. Therefore, the cost of enabling partial context extension is a small, fixed precomputation that does not impact the per-token generation latency.

\section{On Deriving Closed-Form Hyperparameter Formulae}
\label{appendix:closed_form}

A practical extension of this work is to derive closed-form formulae for determining the optimal cutoff dimension index and attention temperature for Partial YaRN akin to the solution provided in YaRN for unimodal LLMs \citep{yarn}, which would eliminate the cost of hyperparameter searches. We leave this for future work due to a key challenge not present in unimodal LLMs: the wide variance in audio representation lengths across different LALMs.

Unlike the relatively consistent token-to-information ratio in text across modern tokenizers, LALMs compress audio into discrete tokens at vastly different rates. For instance, a 30-second audio clip gets compressed into 88 tokens in SALMONN but 750 tokens in Qwen2-Audio. As our sensitivity analysis in Appendix~\ref{Appendix:Hyperparameter Sensitivity of Partial YaRN} shows, this disparity strongly influences the optimal hyperparameter choice, making closed-form formula non-trivial to derive.

\section{Additional Experimental Details}

\subsection{Training and Inference}

\paragraph{Environment.}
All fine-tunings and inferences are conducted using PyTorch \citep{pytorch} and HuggingFace's Transformers \citep{huggingfacetransformerslib} libraries on a single Nvidia H100 GPU. When available, we utilize Flash-Attention 2 \citep{flashattention2} to speed up the processes and reduce VRAM consumptions. For experiments on very long sequence, we use gradient checkpointing to circumvent out-of-memory issue.

\paragraph{Fine-tuning hyperparameters.}
We use LoRA \citep{lora} of rank 8 for fine-tuning through HuggingFace's PEFT library \citep{peft}. With alpha set to 16, and a small LoRA dropout of 5\%. Models are trained with AdamW optimizer \citep{adamw}, learning rate is set to 5e-5, a batch size of 8 samples, and gradient norm clipped to 1.0 \citep{pascanu2013difficulty}.

\subsection{Partial YaRN's Hyperparameter Tuning}
Recall that Partial YaRN has two hyperparameters, cutoff dimension and attention temperature. For the training-free experiment (Section~\ref{sec: Training-free audio context extension}), we jointly sweep the cutoff dimension from $\{56, 48, 40, 32, 24, 16, 8\}$, and the temperature from between 0.5 and 1.6 with a step size of 0.1. See Appendix~\ref{Appendix:Hyperparameter Sensitivity of Partial YaRN} for tuning results and sensitivity.

\begin{figure*}[h]
    \centering
    \includegraphics[width=1.0\textwidth]{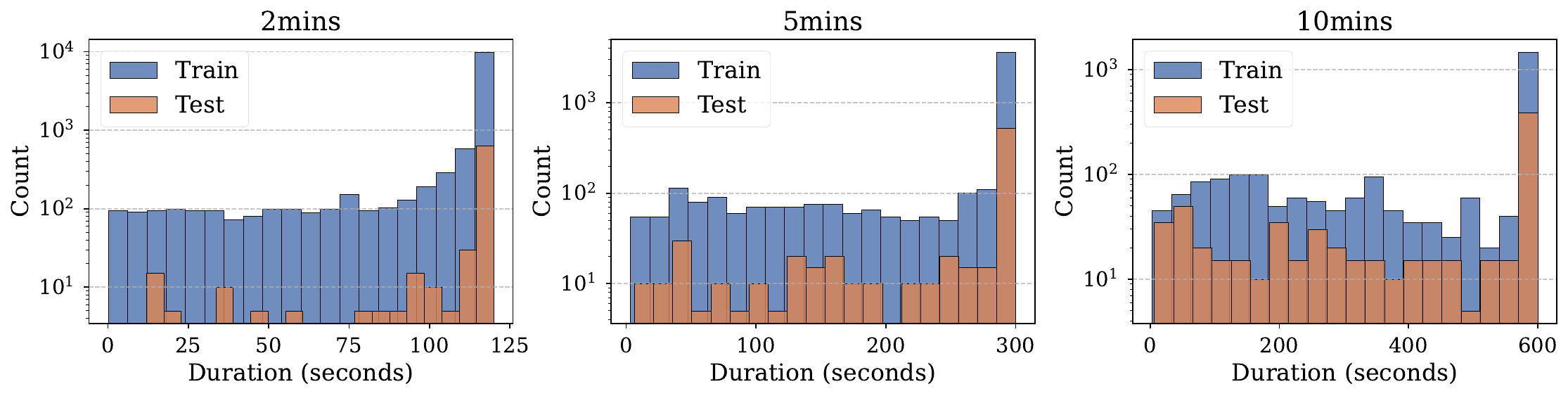}
    \caption{
        Distributions of audio segment lengths in the YODAS2-MCQA dataset, shown with \textit{logarithmic counts} for improved visibility of the long-tail portions.
    }
    
    \label{fig:log_length_distributions}
\end{figure*}

\begin{figure*}[h]
    \centering
    \includegraphics[width=1.0\textwidth]{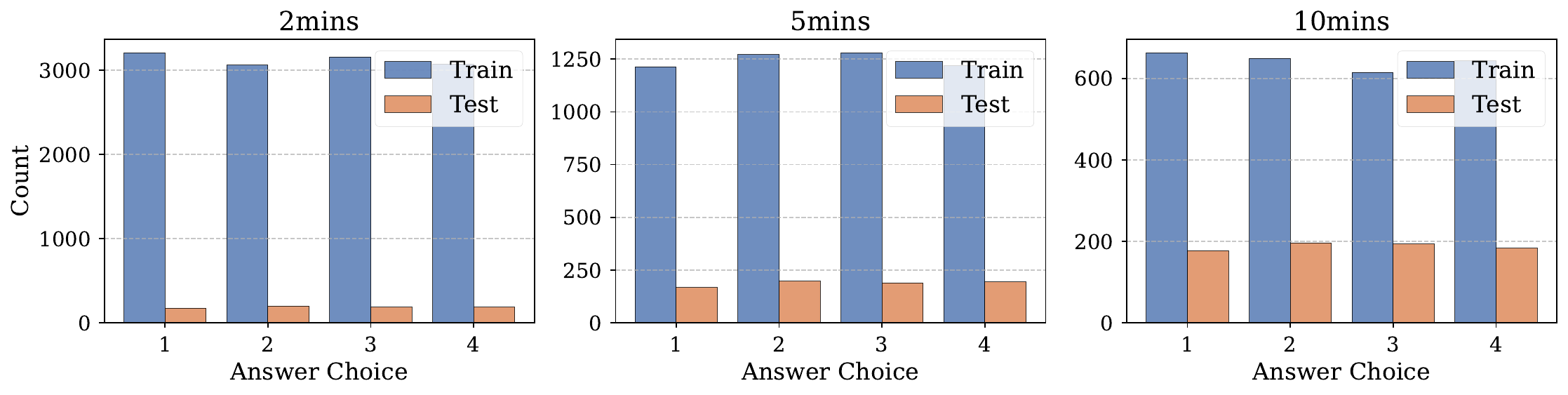}
    \caption{
        Distributions of correct answer choices for all question-answer pairs in the YODAS2-MCQA dataset.
    }
    
    \label{fig:choice_distributions}
\end{figure*}

\begin{figure}[h]
    \begin{tcolorbox}[
        colback=mylightgray,
        colframe=black,
        colbacktitle=mydarkgray,
        coltitle=white,
        fonttitle=\bfseries,
        rounded corners,
        arc=2mm,
        boxrule=1pt,
        boxsep=6pt,
        left=5mm, right=5mm, top=5mm, bottom=5mm,  
        title=\textbf{YODAS2-MCQA inference prompt},
    ]

    Your answer MUST be in the format of "(NUMBER) STATEMENT". For example, if the answer was (4) A pen, you would ONLY output "(4) A pen". Do NOT include any other text. Using information from all the given audios. \texttt{\{question\}}
    
    \end{tcolorbox}
    \caption{Prompt used for inferencing on YODAS2-MCQA.}
\end{figure}

\subsection{Dataset Construction}
\label{appendix: Additional Dataset Details}
Our dataset is derived from a portion of the \textit{en129}\footnote{\url{https://huggingface.co/datasets/espnet/yodas2/tree/main/data/en129}} subset of YODAS2 \citep{li2023yodas}, a large-scale, multilingual corpus of YouTube-derived audio and speech data. The YODAS2 dataset is made available under the CC-BY-3.0 license. We accessed and processed the data using HuggingFace's \textit{Datasets} library \citep{datasetslib}.

To prevent video source leakage in the VLAT experiment, train-test split was performed at the video level; all dialogue segments from a single video were exclusively assigned to either the training set (2mins), or the test sets.

To obtain specific audio length, we chunk the long audio in the dataset to 1, 2, 5, and 10 minutes. Any remaining, incomplete chunk of each long-audio is also retained. Subsequently, we employed Gemini 2.0 Flash \citep{gemini} to generate five unique multiple-choice question-answer (MCQA) pairs for each audio chunk. The complete prompt utilized for this generation process is provided in Figure~\ref{fig: yodas2 mcqa generation prompt}. The resulting dataset, which we have named \textbf{YODAS2-MCQA}, is publicly available at: \texttt{\url{https://huggingface.co/datasets/yophis/yodas2-mcqa}}.

For a more granular analysis of the YODAS2-MCQA dataset, we provide visualizations of the distributions of key characteristics. The distributions of audio segment lengths are presented in Figure~\ref{fig:log_length_distributions}. Additionally, the distribution of correct answers across the multiple-choice options is presented in Figure~\ref{fig:choice_distributions}.

\section{Further Experimental Results}

\subsection{Different VLAT's Virtual Length Sampling Strategies}
\label{Appendix: Different VLAT's Virtual Length Sampling Strategies}
The effectiveness of Virtual Longform Audio Training (VLAT) depends on the strategy used to sample the virtual length during training. For generality, here we define the virtual length as a produce of true audio context length (i.e. 30s) and the \textit{virtual factor}. In our main experiment Section~\ref{sec: Virtual Long Range}, we used a strategy of randomly sampling a factor from the range [1x, 5x, 10x, 15x, 20x, 25x]. To understand the sensitivity of our results to this choice, we compare our default strategy against four alternatives, training a Qwen2-Audio model on each strategy and evaluating on the 10mins audio task.

The virtual factor sampling strategies are:
\begin{enumerate}
    \item \textbf{Default}: Uniform random sampling from $\{ 1.0, 5.0, 10.0, 15.0, 20.0, 25.0 \}$. This cover the audio context length of 30s (1.0x) to 12.5mins (25.0x).
    \item \textbf{Dense Sampling}: Random sampling from a discrete set of 100 uniformly spaced points within the default strategy's range.
    \item \textbf{Very Dense Sampling}: Random sampling from a discrete set of 1000 uniformly spaced points within the default strategy's range.
    \item \textbf{Limited Range}: Uniform random sampling from $\{ 1.0, 5.0, 10.0\}$. This factor set covers audio context length of at most 5mins (10.0x).
    \item \textbf{Fixed High-Factor}: No randomization; the scale factor is always fixed at 20.0x.
\end{enumerate}
We report the result in Table~\ref{tab:vlat_strategies}.

\paragraph{Analysis.} 
The experiment offers several insights into different VLAT sampling strategies: (1) \textit{Fixed High-Factor} (always 20.0x) strategy performs the worst by a large margin, \uline{confirming that exposure to a variety of positional ranges is crucial for robust generalization, rather than training for just a single fixed target length.} (2) \textit{Default} strategy which uses a coarse set of 6 sampling points outperforms both the \textit{Dense} (100 points) and \textit{Very Dense} (1000 points) strategies. This suggests that finegrained sampling of virtual factors provides no additional benefit, and that a relatively small but diverse set of factors is sufficient for training length-robust models.

The most surprising result comes from the \textit{Limited Range} strategy, which achieves the highest performance across all evaluation lengths, despite its virtual lengths not extending beyond the 5mins mark (10.0x factor). While this demonstrates a powerful capacity for extrapolation, the mechanism behind this superior performance is not immediately clear. \uline{It may suggest that randomization on extremely long virtual lengths could introduce instability to the training}, which could in turn explain why Whole YaRN diverged quickly under VLAT framework; or that the specific distribution of factors in this limited set is coincidentally optimal for our test data. We left the more in-depth study of this behavior as future work.

Given this uncertainty, we retained the \textit{Default} strategy for our main experiments in Section~\ref{sec: Virtual Long Range}. Its factors ranges intuitively and completely covers the lengths seen during evaluation, making it a more principled and safer choice.

\begin{table}[h]
    \setlength\tabcolsep{4pt}
    \small
    \centering
    \begin{tabular}{lccc}
        \toprule
        & \multicolumn{3}{c}{\textbf{YODAS2-MCQA}} \\
        \cmidrule(lr){2-4}
        \textbf{Virtual Length Strategy} & {2 mins} & {5 mins} & {10 mins} \\
        \midrule
        Default & \uline{96.87} & \uline{91.91} & \uline{81.73} \\
        100 sampling points & 95.91 & 89.25 & 79.78 \\
        1000 sampling points & 96.18 & 87.69 & 79.60 \\
        Limited range & \textbf{98.00} & \textbf{93.55} & \textbf{89.20} \\
        Fixed to 20.0x & 89.60 & 68.71 & 50.31 \\
        \bottomrule
    \end{tabular}
    \caption{\textbf{VLAT performance with different sampling strategies.} All models are trained on 2mins audio and evaluated with Partial PI inference. The results are averaged across three random seeds.}
    \label{tab:vlat_strategies}
\end{table}

\subsection{Hyperparameter Sensitivity of Partial YaRN}
\label{Appendix:Hyperparameter Sensitivity of Partial YaRN}

Here we analyze the effect and sensitivity of different Partial YaRN's hyperparameter configurations. We report validation accuracies w.r.t. different cutoff dimension indices and attention temperatures, for the Qwen2-Audio and SALMONN models.

\subsubsection{Sensitivity on Qwen2-Audio}

\begin{figure*}[t]
    \centering
    \includegraphics[width=1.0\textwidth]{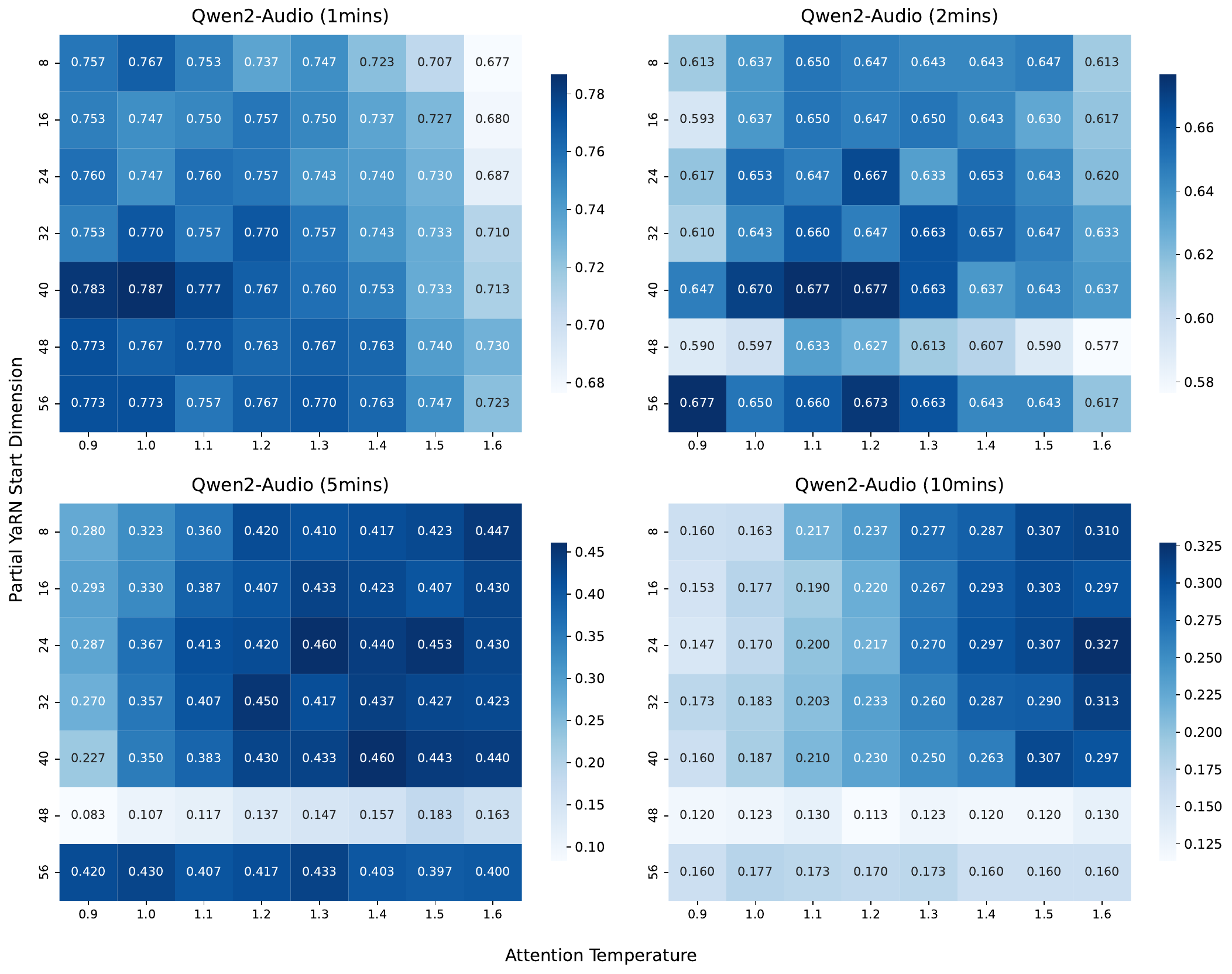}
    \caption{
        \textbf{Hyperparameter Sensitivity of Partial YaRN on Qwen2-Audio (extending the original 30s audio context).}
    }
    
    \label{fig:qwen_hparams}
\end{figure*}

\begin{figure*}[h]
    \centering
    \includegraphics[width=1.0\textwidth]{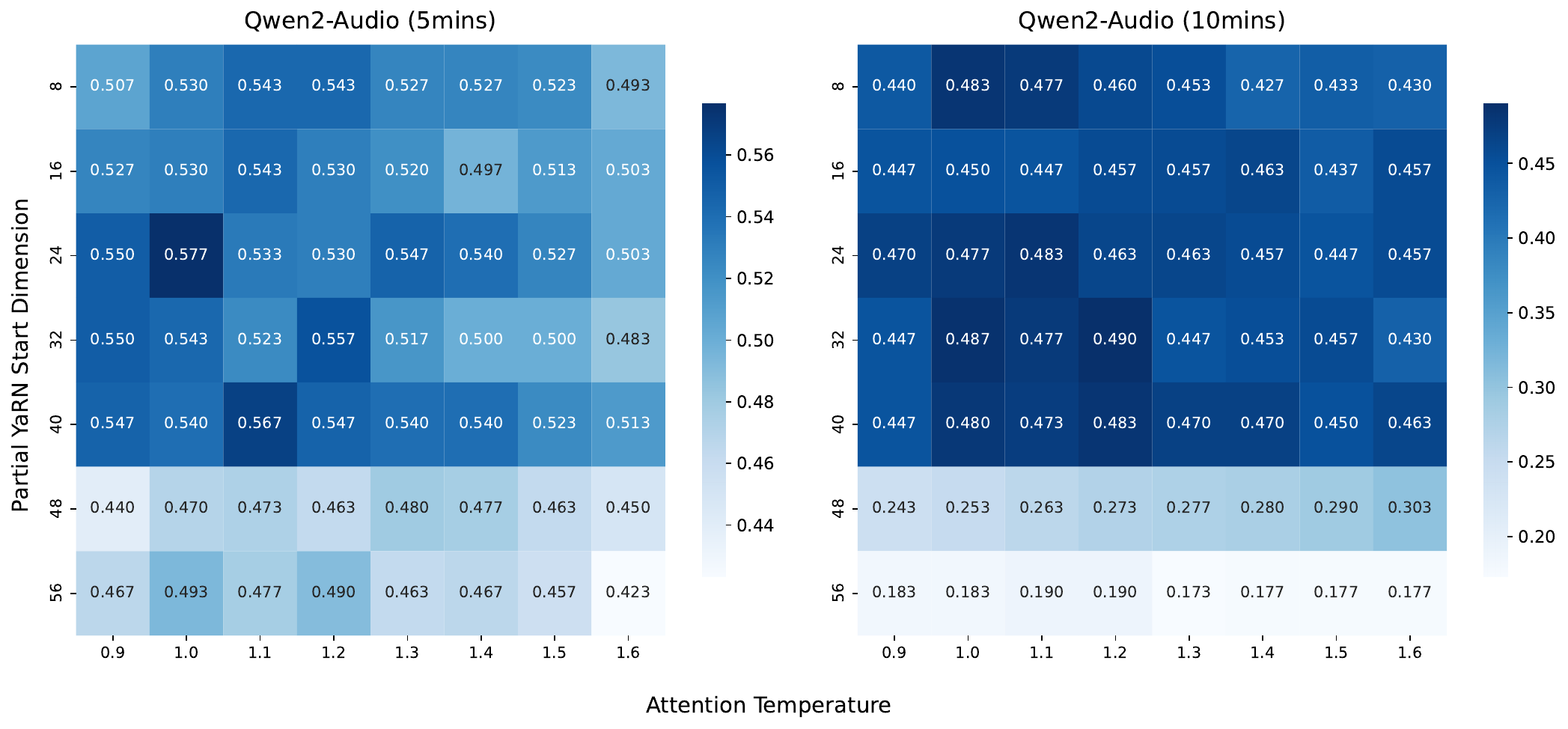}
    \caption{
        \textbf{Hyperparameter Sensitivity of Partial YaRN on Qwen2-Audio (extending the observed 2mins audio context).}
    }
    
    \label{fig:qwen_hparams_stretch_observed}
\end{figure*}

Figure~\ref{fig:qwen_hparams} displays the hyperparameter sensitivity of Partial YaRN on Qwen2-Audio when extending from the original 30s audio context window. Figure~\ref{fig:qwen_hparams_stretch_observed} shows the sensitivity when extending from the observed audio context window of 2mins. 

\paragraph{Sensitivity trend.}
We observe the trend of optimal hyperparameter configurations to move from the bottom left (high cutoff dimension index, temperature close to 1.0) toward the top right of the heatmaps (lower cutoff dimension index, higher temperature) as the audio length increases.

\paragraph{Hyperparameter recommendation.}
This trend informs the following recommendation when using Partial YaRN. For moderate context extensions (e.g., 2x to 4x), it is advisable to start with a higher cutoff dimension index and a temperature value close to 1.0. For longer extensions, a lower cutoff dimension index with a considerably higher temperature is likely to yield better results.

\paragraph{Plausible explanation of this trend.}
First, the attention temperature controls the distribution of attention scores. In very long sequences with thousands of tokens, the standard softmax function can lead to an uneven attention score distribution, such as attending to only a few tokens while ignoring the vast majority of the context. Raising the temperature flattens this distribution, encouraging the model to maintain a broader attention pattern across the entire long audio stream and preventing "attention collapse".

Second, the cutoff dimension index directly controls the "compression pressure" of the position encoding extension. A high cutoff index is expressive enough for shorter extensions while retaining majority of the original, detailed positional information. On the other hand, a lower cutoff index (higher compression) is required to map a much larger range of positions into the smaller familiar audio region.

\subsubsection{Sensitivity on SALMONN}

\begin{figure*}[t]
    \centering
    \includegraphics[width=1.0\textwidth]{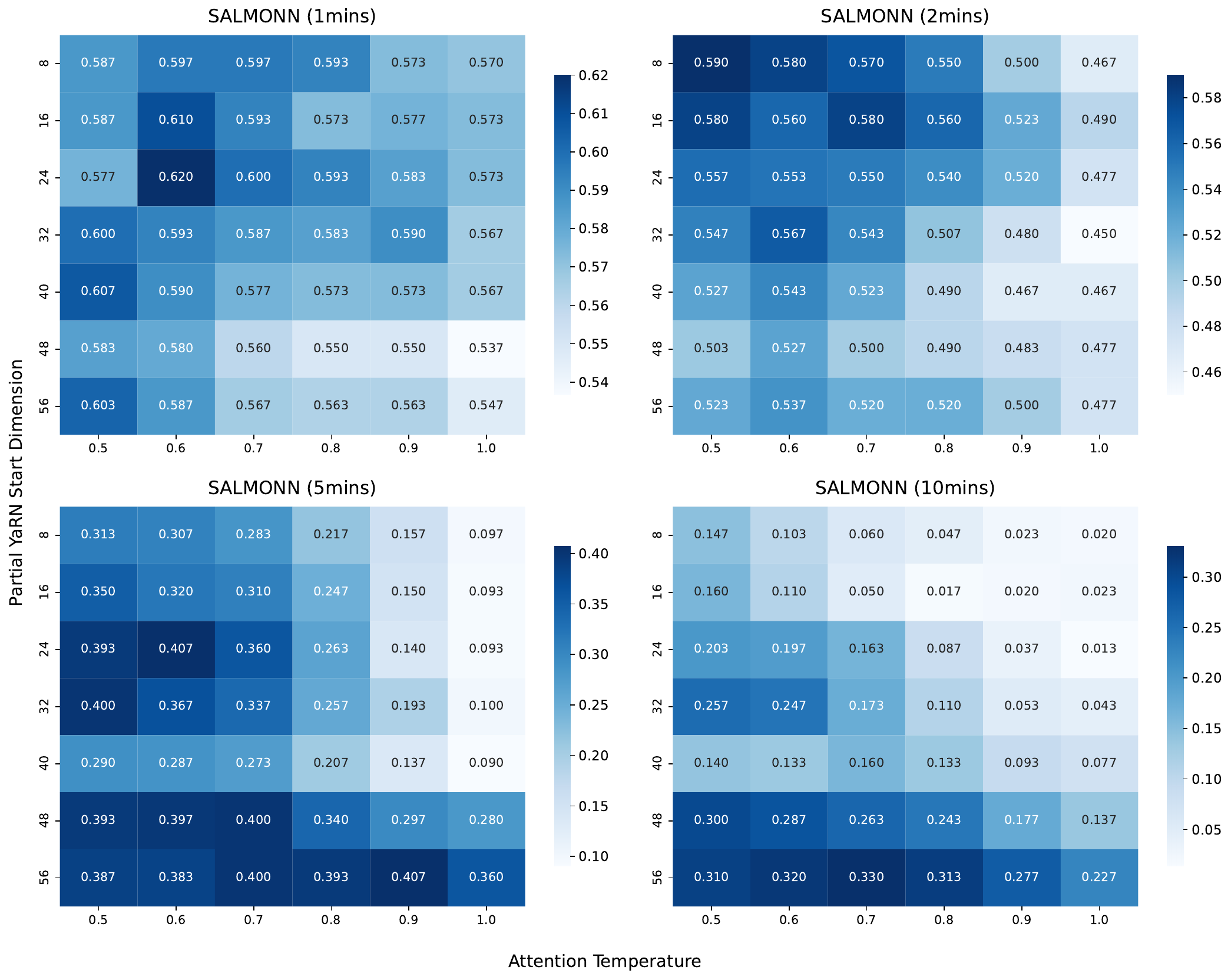}
    \caption{
        \textbf{Hyperparameter Sensitivity of Partial YaRN on SALMONN (extending the original 30s audio context).}
    }
    
    \label{fig:salmonn_hparams}
\end{figure*}

Figure~\ref{fig:salmonn_hparams} displays the hyperparameter sensitivity of Partial YaRN on SALMONN when extending from the original 30s audio context window. 

\paragraph{Sensitivity trend.}
For this model, we observe that the optimal hyperparameter sets are generally contained in the low temperature region (0.5 - 0.7), with the cutoff dimension index moving from low to higher ones as the audio length grows.

\paragraph{Hyperparameter recommendation.}
From the observed trend, it's advisable to set the attention temperature to $0.6 (\pm 0.1)$, adopt a low cutoff index for shorter audio length, and higher index for longer audio inputs.

\paragraph{Plausible explanation of this trend.}
The vast distinction between SALMONN and the previously discussed Qwen2-Audio lies in the their audio representation lengths---SALMONN only uses 88 tokens for every 30s audio vs. Qwen2-Audio's 750 tokens. This, coupled with the model's preference of low attention temperature, suggests that letting the model focuses on a relevant local region outweighs the need to spreading the attention distribution as in Qwen2-Audio.

\begin{figure*}[t]
    \centering
    \includegraphics[width=1.0\textwidth]{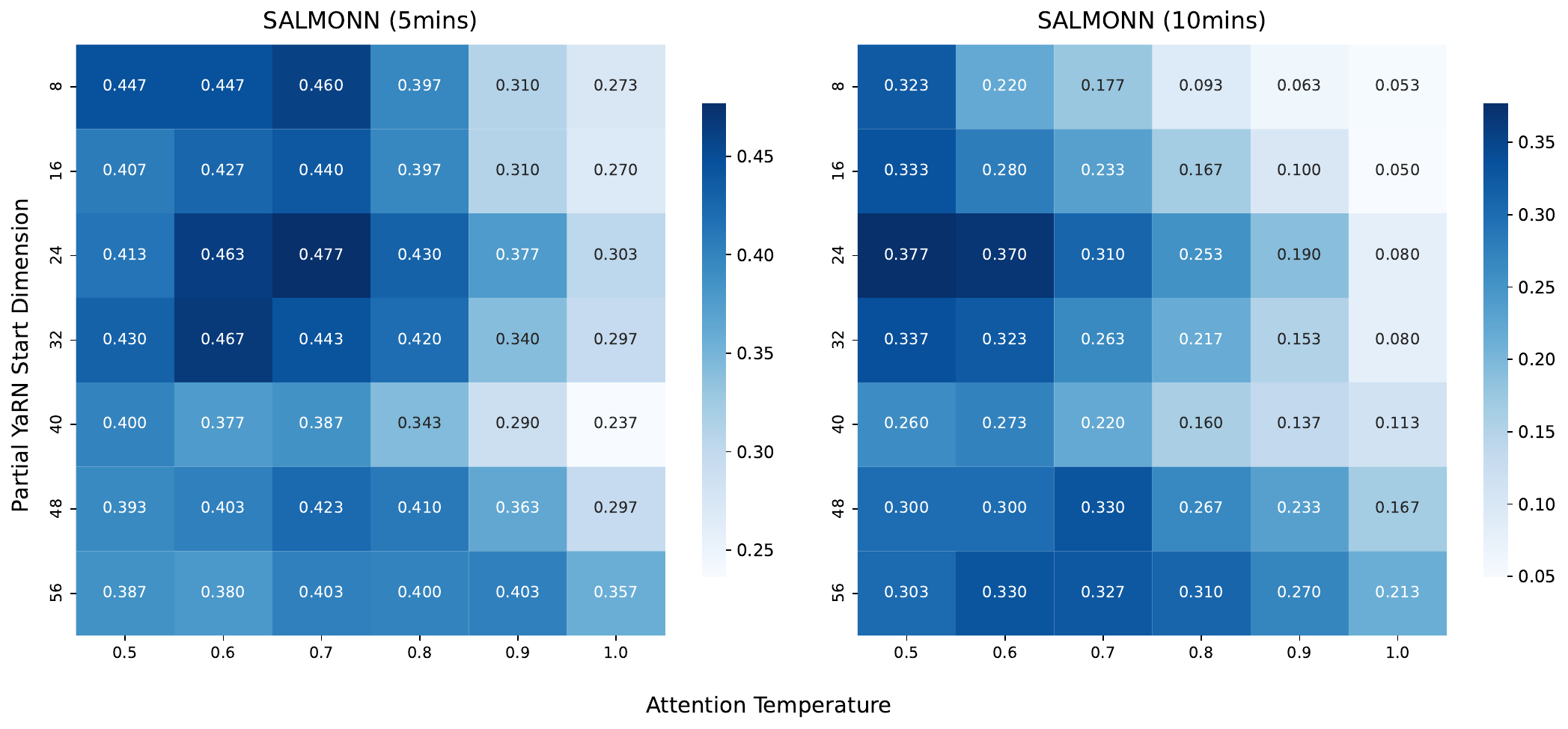}
    \caption{
        \textbf{Hyperparameter Sensitivity of Partial YaRN on SALMONN (extending the observed 2mins audio context).}
    }
    
    \label{fig:salmonn_hparams_stretch_observed}
\end{figure*}

\begin{figure*}[p!]
    \begin{tcolorbox}[
        colback=mylightgray,
        colframe=black,
        colbacktitle=mydarkgray,
        coltitle=white,
        fonttitle=\bfseries,
        rounded corners,
        arc=2mm,
        boxrule=1pt,
        boxsep=6pt,
        left=5mm, right=5mm, top=5mm, bottom=5mm,  
        title=\textbf{YODAS2-MCQA generation prompt},
        fontupper=\scriptsize
    ]

You are an expert in creating high-quality Multiple-Choice Question Answering (MCQA) datasets for speech understanding. Your task is to generate QA pairs based on audio segments.\\

**Input:**

You will be given a segment of an audio recording.\\

**Task:**

For each given transcript, generate **5 distinct 4-Choice Question-Answer (QA) pairs** that each has 1 correct and 3 incorrect choices (so four choices in total), and are factually grounded in the content of the transcript. The questions should cover different aspects of the audio segment, such as:

*   **Who:** Identify speakers or people mentioned.\\
*   **What:** Describe actions, objects, or events.\\
*   **When:** Determine times or dates mentioned.\\
*   **Where:** Locate places or locations.\\
*   **Why/How:** Explain reasons or processes.\\
*   **Details:** Ask for specific details mentioned in the audio segment.\\
*   **Purpose/Intention:** Infer the speaker's intent.\\
*   **Summary:** Condense information from the audio segment.\\
*   **Count:** How many people, men or women in the audio segment.\\
*   **Opinion/Sentiment:** What opinions or sentiment is expressed.\\
*   **Negation:** Understand what is NOT being said or done.\\

**Output:**

Your output should be a JSON formatted string. The JSON should have the following structure:

{
  "response": [
        {"question": "Question 1", "correct": "Correct answer", "incorrect1": "Incorrect answer1", "incorrect2": "Incorrect answer2", "incorrect3": "Incorrect answer3"},\\
        {"question": "Question 2", "correct": "Correct answer", "incorrect1": "Incorrect answer1", "incorrect2": "Incorrect answer2", "incorrect3": "Incorrect answer3"},\\
        {"question": "Question 3", "correct": "Correct answer", "incorrect1": "Incorrect answer1", "incorrect2": "Incorrect answer2", "incorrect3": "Incorrect answer3"},\\
        {"question": "Question 4", "correct": "Correct answer", "incorrect1": "Incorrect answer1", "incorrect2": "Incorrect answer2", "incorrect3": "Incorrect answer3"},\\
        {"question": "Question 5", "correct": "Correct answer", "incorrect1": "Incorrect answer1", "incorrect2": "Incorrect answer2", "incorrect3": "Incorrect answer3"}
  ]
}\\

Important Guidelines:\\
Accuracy: Ensure the answers are directly supported by content of the audio. Do not introduce information that is not present in the audio.\\
Relevance: The questions should be relevant to the content of the audio segment.\\
Diversity: The 5 QA pairs should be diverse in the type of question asked (who, what, when, where, why, how, detail, etc.). Try to use each question type at least once, if applicable to the transcript.\\
Clarity: The questions and answers should be clear, concise, and grammatically correct.\\
Completeness: The answers should fully address the question and provide sufficient information.\\
Conciseness: The answers should be as concise as possible while still being accurate and complete.\\
Coverage: The questions should collectively cover the entire audio segment. Aim for diverse coverage for each question. Prioritize the most salient and important information.\\
Avoid Ambiguity: Ensure the questions are not ambiguous and have a clear, single answer.\\
Plausibility of Incorrect Answers: The incorrect answers should be plausible but wrong, and related to the context of the transcript. Don't make them obviously incorrect or nonsensical. Use information or wording from the transcript to make them believable distractors.\\
JSON Validity: The output must be a valid JSON string. Double-check the formatting and escaping of special characters.\\
Context-Awareness: While each audio segment should be treated independently, the types of questions should logically relate to the context of human conversations. Try to create questions which someone would naturally ask.\\
Answer Length Consistency: Try to make the length of the correct and incorrect answers as similar as possible. This makes it more difficult to guess the answer based on length alone.\\

Example:\\
Input Transcript: "Okay, so the meeting is scheduled for next Tuesday at 2 PM in Conference Room A. We'll be discussing the Q3 marketing report and the proposed budget changes."\\
Output:

{
  "response": [
    {
      "question": "When is the meeting scheduled?",
      "correct": "Next Tuesday at 2 PM",
      "incorrect1": "This Friday at 10 AM",
      "incorrect2": "Next Wednesday at 3 PM",
      "incorrect3": "Today at 4 PM"
    },
    {
      "question": "Where will the meeting take place?",
      "correct": "Conference Room A",
      "incorrect1": "Conference Room B",
      "incorrect2": "The Main Auditorium",
      "incorrect3": "The Executive Boardroom"
    },
    {
      "question": "What marketing report will be discussed?",
      "correct": "The Q3 marketing report",
      "incorrect1": "The Q2 marketing report",
      "incorrect2": "The annual marketing report",
      "incorrect3": "The preliminary marketing report"
    },
    {
      "question": "Besides the marketing report, what else will be discussed?",
      "correct": "The proposed budget changes",
      "incorrect1": "The sales figures",
      "incorrect2": "The employee satisfaction survey",
      "incorrect3": "The new product launch"
    },
    {
      "question": "What is the purpose of the meeting?",
      "correct": "To discuss the Q3 marketing report and budget changes",
      "incorrect1": "To plan the annual company picnic",
      "incorrect2": "To interview new job candidates",
      "incorrect3": "To celebrate the company's anniversary"
    }
  ]
}\\

Now, generate 5 QA pairs in the specified JSON format for the input audio:

    \end{tcolorbox}
    
    \centering
    \caption{Prompt for generating multiple-choice question answering pairs from YODAS2.}
    \label{fig: yodas2 mcqa generation prompt}
\end{figure*}

\end{document}